\documentclass{article}

     \PassOptionsToPackage{numbers, compress}{natbib}

     \usepackage[final]{neurips_2020}

\usepackage[utf8]{inputenc} 
\usepackage[T1]{fontenc}    
\usepackage{hyperref}       
\usepackage{url}            
\usepackage{booktabs}       
\usepackage{amsfonts}       
\usepackage{nicefrac}       
\usepackage{microtype}      
\usepackage{amsmath}
\usepackage{graphicx}
\usepackage{dsfont}
\usepackage{subfigure}
\usepackage{stmaryrd}
\usepackage{float}
\usepackage{wrapfig}
\usepackage{gensymb}
\usepackage{graphicx}
\usepackage{comment}
\usepackage{amsmath,amssymb} 
\usepackage{color}
\usepackage{epsfig}
\usepackage{array}
\usepackage{multirow}
\usepackage{amsfonts}
\usepackage{caption}
\usepackage{wrapfig}
\usepackage{verbatim}
\usepackage{float}
\usepackage{enumitem}
\usepackage{booktabs}
\usepackage{lipsum}
\usepackage{subfigure}
\usepackage{tabulary,overpic,xcolor}

\newcommand{\ie}{\textit{i}.\textit{e}.}
\newcommand{\eg}{\textit{e}.\textit{g}.}

\newcommand{\xx}{\mathbf{x}}
\newcommand{\XX}{\mathbf{X}}
\newcommand{\yy}{\mathbf{y}}
\newcommand{\RR}{\mathbb{R}}

\newcommand{\NN}{\mathcal{N}}

\newcommand{\zz}{\mathbf{z}}

\newcommand{\ZZ}{\mathbf{Z}}

\newcommand{\s}{\mathbf{s}}

\newcommand{\bd}[1]{\textbf{#1}}

\title{Structure-Regularized Attention for Deformable Object Representation}

\author{
  Shenao Zhang \\
  Georgia Institute of Technology \\
  \texttt{shenao@gatech.edu}\\
    \And
  Li Shen\\
  Tencent AI Lab \\
  \texttt{lshen.lsh@gmail.com} \\
      \AND
  Zhifeng Li  \\
  Tencent AI Lab \\
  \texttt{michaelzfli@tencent.com} \\
      \And
  Wei Liu \\
  Tencent AI Lab \\
  \texttt{wl2223@columbia.edu} \\
}

\begin{document}

\maketitle

\begin{abstract}
Capturing contextual dependencies has proven useful to improve the representational power of deep neural networks. 
Recent approaches that focus on modeling global context, such as self-attention and non-local operation, achieve this goal by enabling unconstrained pairwise interactions between elements. 
In this work, we consider learning representations for deformable objects which can benefit from context exploitation by modeling the structural dependencies that the data intrinsically possesses. To this end, we provide a novel structure-regularized attention mechanism, which formalizes feature interaction as structural factorization through the use of a pair of light-weight operations. The instantiated building blocks can be directly incorporated into modern convolutional neural networks, to boost the representational power in an efficient manner. Comprehensive studies on multiple tasks and empirical comparisons with modern attention mechanisms demonstrate the gains brought by our method in terms of both performance and model complexity. We further investigate its effect on feature representations, showing that our trained models can capture diversified representations characterizing object parts without resorting to extra supervision.
\end{abstract}

\section{Introduction}
\label{sec:intro}
Attention is capable of learning to focus on the most informative or relevant components of input and has proven to be an effective approach for boosting the performance of neural networks on a wide range of tasks \cite{itti1998model,itti_2001,vaswani_nips2017,hu2018squeeze,wang2018non}. Self-attention \cite{vaswani_nips2017} is an instantiation of attention which weights the context elements by leveraging pairwise dependencies between the representations of query and every contextual element. The ability of exploiting the entire context with variable length has allowed it to be successfully integrated into the encoder-decoder framework for sequence processing.  
\cite{wang2018non} interprets it as non-local means \cite{buades2011non}, and adapts it to convolutional neural networks.
However, it is computationally expensive where the complexity is quadratic with respect to input length (e.g., spatial dimensions for image and spatial-temporal dimensions for video sequence). The approaches capture long-range association by allowing each node (e.g., a pixel on feature maps) to attend over every other positions, forming a complete and unconstrained graph which may be intractable for extracting informative patterns in practice. Relative position embedding has proven useful in alleviating the issue \cite{bello2019attention,shaw2018self} but the structural information specific to tasks is not effectively exploited.

\begin{figure}[t]
\begin{center}
\includegraphics[width=0.7\textwidth]{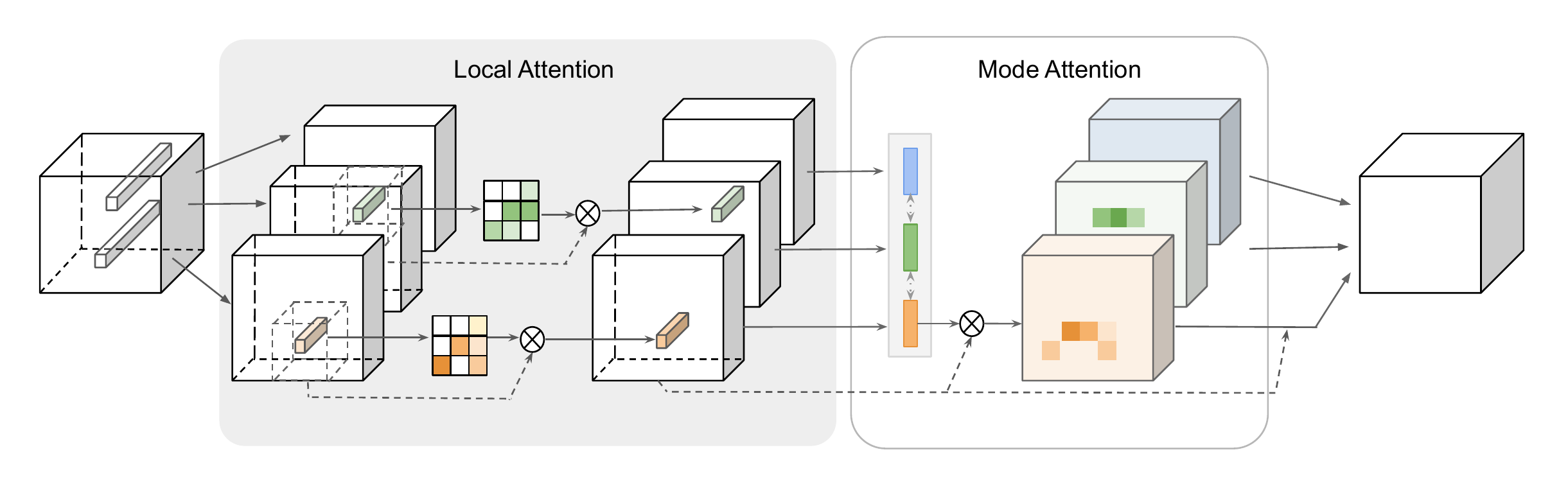}
\end{center}
\vspace{-4.5mm}
\caption{The illustration of the Structure-Regularized Attention Block.}
\label{fig:block}
\end{figure}

In this work, we aim to address the visual tasks of representing naturally deformable objects \cite{felzenszwalb2008discriminatively,thewlis2017unsupervised}, such as face or bicycle, which have appearance and shape variance when encountering different viewing conditions or deformations, and may require or highly benefit from using the structure prior data intrinsically has. Inspired from subspace algorithms \cite{parsons2004subspace, tipping1999mixtures, vidal2005generalized}, the design is built upon the hypothesis of data factorizability, i.e., projecting the data into multiple feature subspaces, defined as modes, which are expected to be more compact and typically represent certain components of objects. A set of parameterized transformations project nodes into multiple modes and capture correlation between nodes and modes, aiming to learn discriminative representations by effectively modeling structural dependencies. We achieve the goal by introducing a novel attention module, which is termed "Structure-Regularized Attention" (StRAttention), formalized as the composition of two-level operations, namely local and mode attentions (in Fig.~\ref{fig:block}). The local attention functions as spatial expansion on local regions.
The higher-level contextual information can be accessed through the mode attention, allowing diversified contextual information to be distributed. The mechanism enables each node to attend to (theoretically) global context in a structural manner. Code is available at \url{https://github.com/shenao-zhang/StRA}.

\section{Method}
\label{sec:alg}
\vspace{-2mm}
We will use "pixel" and "node" interchangeably, and "mode" and "group" interchangeably in the following descriptions. Formally, let $\XX \in \RR^{H \times W \times C}$ denote an input, e.g., feature maps of an image, with spatial dimensions $H \times W$ and channels $C$, and $\xx_{i}$ denote the feature on  pixel $x_{i}$, where $i \in \NN_G \equiv \{1, \cdots, HW\}$, and $\NN_G$ denotes the set of spatial dimensions.

The contextual feature is captured by virtue of the transformation $f: \RR^{C}\times \RR^{C} \rightarrow [0,1]$ \cite{vaswani_nips2017,wang2018non},
$\yy_{i} = \sum_{j\in \NN_G} f(\xx_{i}, \xx_{j})u(\xx_{j})$,
where $u$ represents the unitary transformation on a single node
and $f$ captures the pairwise relation between nodes within global context. Here $f$ forms a complete graph in which each node can attend to every other node. It brings about the challenge of a quadratic computational complexity and memory overhead with respect to the size of $\NN_G$ \cite{huang2019ccnet,li2019expectation}.

The use of hierarchical structure is believed to play a critical role in capturing the statistics in images independent of learnable parameters \cite{ulyanov2018deep}. In reality, most data can be assumed to live on low dimensional manifolds. 
To this end, we formalize the problem as a form of structural factorization. We want to learn a set of transformations to project data onto multiple diversified subspaces, $\Phi:=\{{\rm\Phi}_g\}_{g=1}^G$, where ${\rm\Phi}_g: \mathcal{X} \rightarrow \mathcal{S}_g$ corresponds to the projection from the universal feature space (i.e., input feature maps) onto the $g$-th subspace which we call "mode" here. The corresponding output of node $x_i$ is represented as $\s^g_i$. Each mode is expected to represent a certain factor the data consists of (e.g., discriminative parts), denoted by modal vectors $\ZZ = \{\zz_g\}_{g=1}^G$ for input $\mathbf{X}$. The modal vectors are generated by integrating the projections through the function $\xi_g: \mathbf{S}_g \mapsto \zz_g$ where $\mathbf{S}_g = \{\mathbf{s}_i^g\}_{i\in \mathcal{N}_G}$. Let $r_{ig}\in [0,1] $ indicate the matching degree of node $x_i$ with respect to the $g$-th mode, which we term {\it attention coefficients}. Then the context for node $x_i$ is formulated as a combination of the information derived from each mode $\yy_{i} := \bigcup \yy_i^g$, and 
\begin{equation}
\label{eq:ma}
\yy_{i}^g = r_{ig} \cdot \zz_g, \quad r_{ig}= {\gamma}(\s^g_i, \zz_g),
\end{equation}
where the information between modes can be further correlated through a function $\rho: \mathcal{Z}\rightarrow\mathcal{Z}$ that captures the relation between modal vectors and propagates such higher-level context to each node. 

\vspace{-1mm}
\textbf{Local Attention.} The transformation onto the $g$-th subspace ${\rm\Phi}_g$ can be implemented by convolutions. The index $g$ is omitted for simplification. We propose an alternative which is defined as,
\begin{equation}
\label{eq:la_2}
\s_{i}=\sum_{j\in \NN_K(i)} a_{ij} u(\xx_j), \ a_{ij}= \sigma_m\left(\omega(\xx_i)_{j} + \nu(\xx_j)\right),
\end{equation}
where $\sigma_m$ denotes the softmax function, $\omega: \RR^C \rightarrow \RR^{K\cdot K}$ and $\nu: \RR^C \rightarrow \RR$. The affinity matrix $A_i= \{a_{ij}\}_{j\in\NN_K(i)} \in [0, 1]^{K\times K}$ is expected to generate a proper {\it data-dependent} local softmask on the $K\times K$ neighbourhood of each node $x_i$ for local context aggregation.

\textbf{Mode Attention.}
\label{sec:ma}
A deformable object can be effectively described by a combination of representations towards different parts \cite{felzenszwalb2008discriminatively}. 
We expect each mode responsible for the feature distribution of one distinct component whose intrinsic properties are described by modal vectors, \ie, $\xi_g$ is realized by mean features (averaging over the local attention output $\mathbf{S}_g$) or centroid features (averaging over representative nodes). The attention coefficient $r_{ig}$ in (\ref{eq:ma}) is then measured by inner product between the corresponding feature vector for node $x_i$ and the modal vector:
\begin{equation}
\label{eq:sig}
r_{ig} = \gamma(\s_i^g, \zz_g) = \sigma(\langle\s_i^g, \zz_g \rangle).
\end{equation} 
$\sigma$ denotes the gating, which can be defined as either softmax or sigmoid function, representing that the relation is modelled in a mutually exclusive or independently manner. Both forms can achieve the goal that nodes will share some context induced by modes to enhance their desired representations.
Mode interaction $\rho: \mathcal{Z}\rightarrow\mathcal{Z}$ can be conveniently achieved by,
$\zz'_{g} = \sum_{j=1}^G \sigma_m(\langle\zz_g, \zz_j \rangle)\cdot\zz_j$. The updated $\ZZ$ of across-mode interactions can substitute that in (\ref{eq:sig}), which is complementary (added) to the output of local attention. Detailed implementation and module schema are shown in the Appendix.
\vspace{-1mm}
\textit{Discussion.} The design of correlating nodes to multiple modes is related to soft-clustering and mixture models \cite{mclachlan1988mixture} which learn clusters by updating central vectors and node assignments iteratively through Expectation-Maximization algorithm \cite{dempster1977maximum}.  Such an iterative process is substituted by forward and backward propagation in the framework where the associated parameters are learned by gradient descent. During inference the modal vectors and the attention coefficients are computed once, which is more efficient and suitable for neural network paradigms.

\section{Experiments}
\label{sec:exp}
\vspace{-2mm}
To validate the effectiveness of the proposed StRA, we conduct experiments on two types of widely studied deformable objects: human body and human face. We will focus on three tasks: person re-identification (ReID), face recognition and facial expression recognition.
\vspace{-1mm}
\subsection{Human Body}
\vspace{-1mm}
We evaluate the method mainly on the Market1501 dataset\cite{zheng2015scalable} of the person ReID task.
\begin{figure}[t]
\begin{minipage}{0.45\textwidth}
\centering
\captionof{table}{Comparison on Market1501.}
\renewcommand\arraystretch{1.05}
\scalebox{0.8}{
\begin{tabular}
{l p{0.6cm}<{\centering} p{0.7cm}<{\centering}p{0.8cm}<{\centering}}
\toprule
Network & mAP  & Rank1 & FLOPs\\
\midrule
ResNet50 \cite{he_cvpr2016resnet} & 77.1 & 90.6 & 4.05G \\
SASA \cite{ramachandran2019stand} & 79.5 & 92.3  &3.19G\\
SENet \cite{hu2018squeeze} & 80.7 & 93.3  &4.49G\\
Non-local \cite{wang2018non} & 80.2 & 91.9 &7.28G  \\
ResNet50\_StRA & \bd{84.1} & \bd{93.8} &\bd{3.17G} \\
\midrule
mnetv2 \cite{sandler2018mobilenetv2} & 71.7 & 88.7&\bd{370M}   \\
mnetv2\_StRA  & 74.2 & 89.3 & \bd{370M}  \\
mnetv2(1.4) \cite{sandler2018mobilenetv2} & 72.5 & 89.0 & 680M  \\
mnetv2(1.4)\_StRA  & \bd{74.6} & \bd{89.9} & 720M \\
\midrule
\end{tabular}
\label{tab:compare}}
\end{minipage}
\hspace{5mm}
\begin{minipage}{0.45\textwidth}
\includegraphics[scale=0.28]{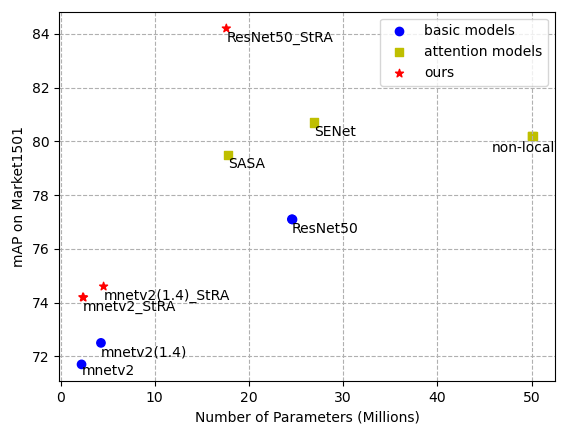}
\centering
\vspace{-2mm}
\caption{Model size vs mAP.}
\label{fig:compare}
\end{minipage}
\end{figure}

Model comparison is conducted on two widely used backbone architectures ResNet \cite{he_cvpr2016resnet} and MobileNetV2 \cite{sandler2018mobilenetv2} in terms of both performance and model complexity. The results in Table \ref{tab:compare} show that our method outperforms the baseline and other attention networks by a large margin on both metrics with the highest efficiency (i.e., Flops). The comparison of parameter sizes (in Fig.\ref{fig:compare}) shows that our method achieves the best trade-off between performance and model complexity on this task. 

\subsection{Face}
%
%

\begin{wraptable}{r}{9cm}
\vspace{-0.6cm}
\caption{Scalability on face recognition (performance \%).}
\label{table:face}
\centering
\vspace{-2mm}
\resizebox{0.65\textwidth}{!}{
\renewcommand\arraystretch{1.15}
\begin{tabular}{c c c| ccccc}
\toprule
\multicolumn{2}{c|}{\multirow{1}*{Dataset}}  & Network & LFW & CFP-FP & CPLFW & CALFW  & AgeDB-30 \\
\midrule
	\multicolumn{2}{c|}{\multirow{2}*{Medium}}  & ResNet50 & 99.1& 94.4& 82.0& 89.1& 93.1 \\
	\cline{3-8}
	\multicolumn{2}{c|}{~} & ResNet50-StRA & 99.1& 95.0& 82.4& 89.5& 93.3 \\
\midrule
    \multicolumn{2}{c|}{\multirow{2}*{Large}} & ResNet50 & 99.5& 97.3& 87.2& 89.9& 93.9 \\
    \cline{3-8}
    \multicolumn{2}{c|}{~} & ResNet50-StRA & 99.7 & 97.5 & 88.0& 91.8&  94.4 \\ 
\bottomrule
\end{tabular}}
\vspace{-0.2cm}
\end{wraptable}

\vspace{-1mm}
\textbf{Face Recognition.} Challenges of face recognition may come from various factors, e.g., variations in pose, expression and illumination. We conduct the experiments to assess the scalability of the method, based on standard classification loss, i.e., softmax loss, for training.
The attention modules are used at the last stage of the architecture. The results in Table \ref{table:face} show that the scalability of the method, i.e., it can consistently enhance the representational power of networks benefiting from increased dataset scale. 

\textbf{Facial Expression Recognition.} Facial expressions explicitly correspond to the deformation of discriminative part/landmarks \cite{cohn2007observer}. With ResNet as backbone, our model can achieve $73.2\%$ accuracy on test set and $71.5\%$ on the public validation set, outperforming the baseline performance, $71.5\%$ and $69.9\%$, by a large margin ($1.7\%$ and $1.6\%$) respectively.

\section{Interpretation and Discussion} 
\label{sec:interp}
\begin{wrapfigure}{r}{7cm}
\centering
\vspace{-3mm}
\subfigure{\includegraphics[width=0.5\textwidth]{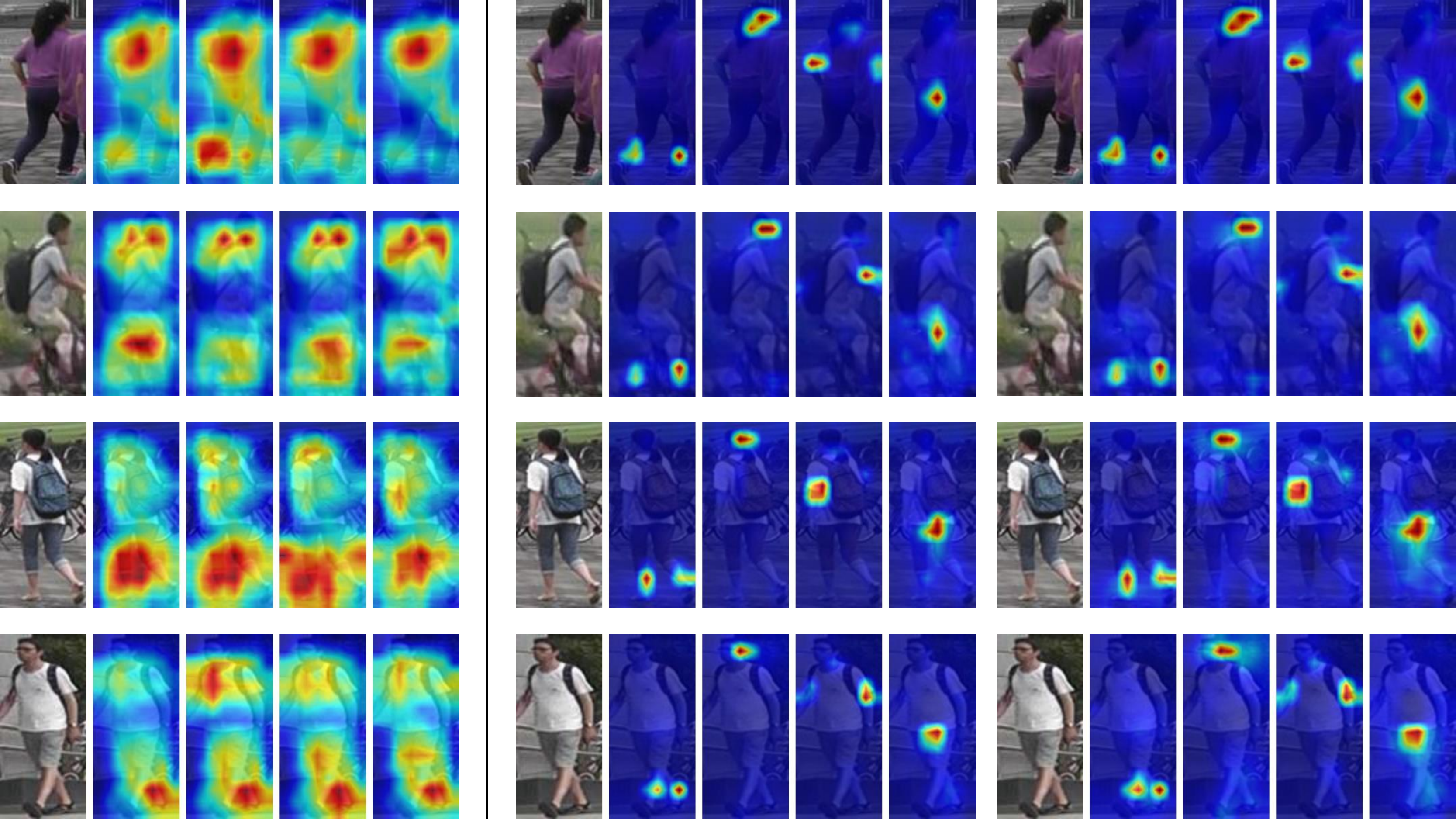}}
\caption{Visualization for the activations of Local Attention variant, attention coefficients in \textit{Mode Attention} and the module outputs.}
\label{fig:vis}
\vspace{-1mm}
\end{wrapfigure}

\paragraph{Activation of structure-distributed representations.}\label{sec:visualization} 
We present the examples of activation visualization (i.e., pixel-wise magnitude on feature maps) for the four modes at the stage 5-1 of ResNet50\_StRA, 
and compare three types of activations in the Fig.~\ref{fig:vis}, \ie, the output of local attention variant running only with local attention, the attention coefficients derived from the mode attention and the final output of the module. The difference between the heatmaps generated by local attention only variant is marginal. Although the multi-head transformations are assumed to detect distinct patterns, the diversity between groups is still difficult to achieve in practice. In contrast, incorporating the regularization of the mode attention unit can diversify feature learning on different groups and encourage exciting features corresponding to discriminative parts of objects, realizing the learning of effective structure-distributed representations.

\begin{figure}[H]
\centering
\subfigure{
    \begin{minipage}[t]{0.22\linewidth}
        \centering
        \includegraphics[width=1.2in]{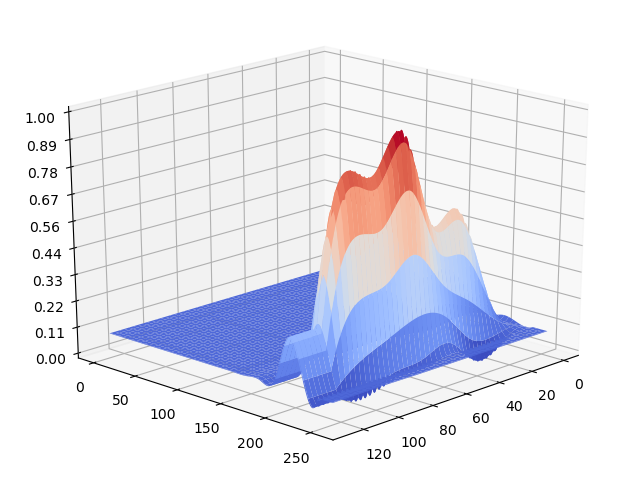}\\
    \end{minipage}%
}%
\subfigure{
    \begin{minipage}[t]{0.22\linewidth}
        \centering
        \includegraphics[width=1.2in]{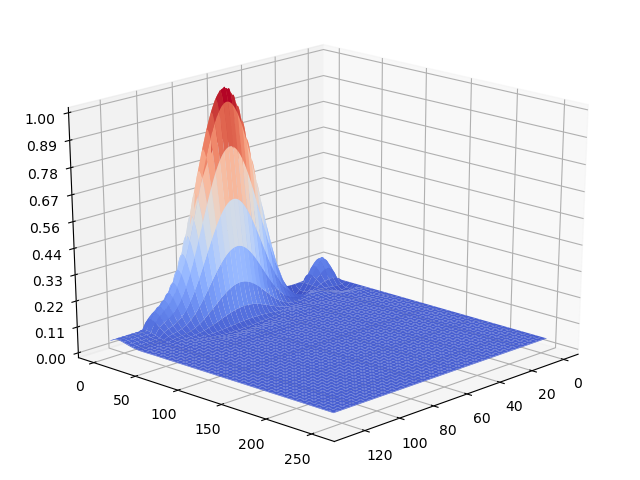}\\
    \end{minipage}%
}%
 \subfigure{
    \begin{minipage}[t]{0.22\linewidth}
        \centering
        \includegraphics[width=1.2in]{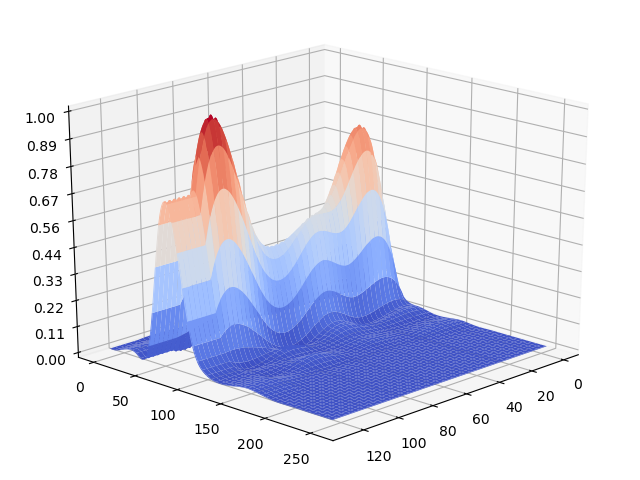}\\
    \end{minipage}%
}%
 \subfigure{
    \begin{minipage}[t]{0.22\linewidth}
        \centering
        \includegraphics[width=1.2in]{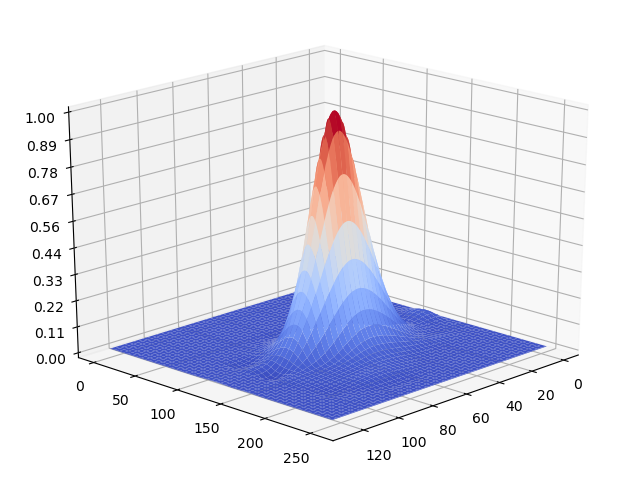}\\
    \end{minipage}%
}%
\centering
\vspace{-2mm}
\caption{Spatial distributions of high activations (i.e., attention coefficients) on the four modes. Higher peaks indicate that more samples are focusing on the corresponding locations.}
\label{fig:3d}
\end{figure}

We also provide the spatial distribution of the four modes (\ie, statistics of attention coefficients with respect to spatial locations across a set of samples) in Fig.~\ref{fig:3d}, showing that nodes (pixels) tend to be projected and highly correlated with certain modes. It demonstrates that such structural regularization mechanism encourages capturing structure-distributed features for deformable objects in a factorized manner, which potentially provides interpretable features.

\section{Conclusion}
In this work we introduced a novel attention module which can effectively capture the long-range dependency for deformable objects through the use of structural factorization on data. The comprised components, i.e., local attention and mode attention, are complementary for capturing the informative patterns and the combination is capable of improving the discriminative power of models. 

The proposed mechanism encourages learning structure-distributed representations which are realized by regularizing information flow conditioned on feature space factorization. The structure prior is assumed to be spatial factorization in the work, where part components are learned in an unsupervised manner (without the need of extra supervision). It would be interesting to generalize to disentangle factors (e.g., describing the factors of age and emotions for face perception) which would widely benefit the representation learning for generative models.
\section{Broader Impact}
\label{broader}
The work presented an insight that representation learning for deformable objects can strongly benefit from exploiting the prior knowledge the data consists of, though recent progress on general network architectures has shown to be transferable to this kind of data. The experiments are conducted on the tasks related to face and human bodies, as they are typical examples of deformable objects. Such tasks may incur some concerns on privacy. The contribution of properly modeling structural dependencies is not confined to such certain task.

We hope that the work can encourage research on network architectures and representation learning for better modeling structural dependencies which will potentially facilitate research on disentangling and interpretable features. The effectiveness of the method may be obstructed by the given subspace number (analogous to subspace number for subspace segmentation methods), especially when the number is difficult to estimate in advance or needs gradually increase during training, while the issue may be addressed from the direction of incorporating network evolution in a nonparametric manner. 

\appendix
\section{Module Implementation}

\begin{figure}[!htb]
\centering
\includegraphics[width=1.0\textwidth]{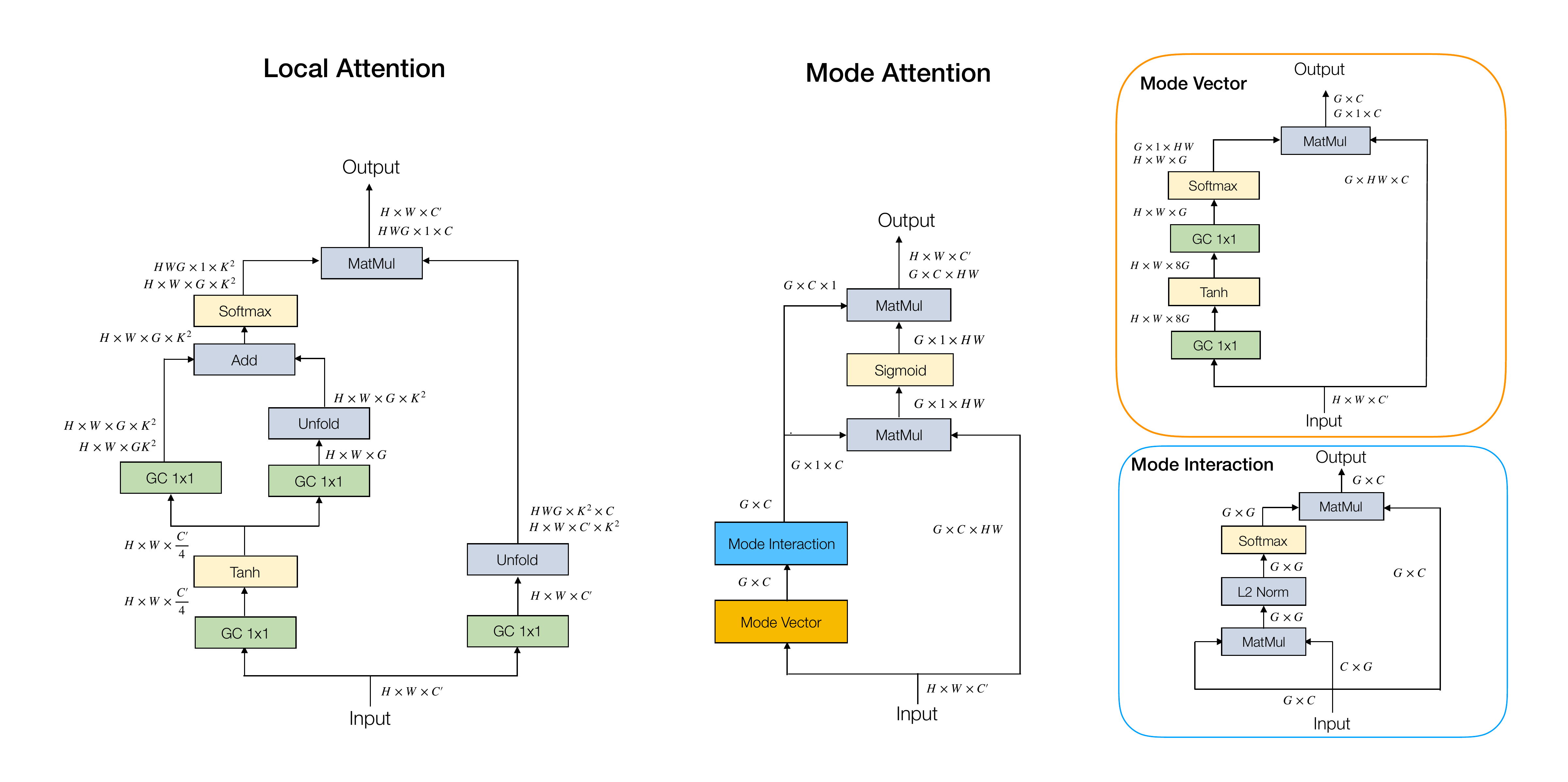}
\caption{The schema of an StRAttention module. \textbf{Left:} \textit{Local Attention} operation (Eq.~\ref{eq:la_2}). The two branches performing with ``Add" operation correspond to the transformation $\omega$ and $\nu$, respectively. The normalized local softmasks (i.e., affinity matrix) multiply with the output of transformation $u$ and produce the operation output. \textbf{Right:} \textit{Mode Attention} operation, which is comprised of mode vector unit and mode interaction unit. G modal vectors are generated by respective normalized spatial masks for each mode. The production of modal vectors and node representations passes through the sigmoid gating function which produces the attention coefficient in Eq.~\ref{eq:sig}. GC denotes group convolutions with the group number set to $G$. 
$C'$ denotes channel dimension, and $C'= C\cdot G$. $H$ and $W$ denote spatial dimensions. $K$ is the local neighborhood size. Batch Normalization \cite{ioffe_icml2015bn} is used after group convolutions by default. The implementation may also require reshaping or permuting operations, which are not explicitly illustrated in this figure.}
\label{fig:module}
\end{figure}

The Structure-Regularized Attention (StRAttention) block is comprised of two operations, \textit{local attention} and \textit{mode attention}. The schema of the two operations, local attention, and mode attention, is shown in Fig~\ref{fig:module}. The input of the local attention operation is produced by a $1\times 1$ convolution and the output of the module is processed by another $1\times 1$ convolution when instantiating it as the drop-in replacement of a bottleneck residual block. As shown in Fig~\ref{fig:block}, input feature maps are dealt with local attention and then fed into mode attention. The outputs of the two operations are added through the use of skip connection. 

The strategy of generating modal vectors is of importance for aggregating the information within modes. In this work the modes are expected to represent spatial structural factorization (e.g., parts or body landmarks), which is modelled in an unsupervised manner.  $1\times 1$ group convolutions equipped with softmax function generate the normalized spatial mask $\mathbf{M}^g \in [0, 1]^{H\times W}$ for each one of $G$ modes. The modal vector is then computed by weighted summation over the output of local attention $\mathbf{S}_g$. Each mode is consequentially represented by the most representative nodes. We impose a diversity regularization on the training loss \cite{thewlis2017unsupervised} to encourage modes to detect disjoint positions, forming a soft constraint for modeling diversified latent factors. The term is formulated as $\mathcal{L}_d= G -\sum\limits_{ij} \max\limits_{g=1,...,G} M_{ij}^g$, which is non-negative and the minimum (i.e., zero value) can be only achieved if disjoint positions are activated with the value $1$.

\section{Experimental Setup}
\subsection{Person ReID}
\textbf{Database:} Market1501 \cite{zheng2015scalable} contains $32,668$ images from $1,501$ identities whose samples are captured under 6 camera viewpoints. $12,936$ images from $751$ identities are used for training and the left images (including $3,368$ query images and $19,732$ gallery images of $750$ persons) are used for testing. 

\textbf{Configuration:} We conduct all the experiments in the single-query setting without a re-ranking algorithm. The results are reported on the cropped images based on detection boxes. We report performance based on two measures: Cumulative matching characteristics (CMC) rank-1 accuracy and mean average precision (mAP). Post-processing (\eg, re-ranking and multi-query fusion) is not applied for all the experiments.

ResNet-50 is used as the backbone architecture, where the last spatial downsampling operation is removed following conventional settings \cite{sun2018beyond,miao2019pose,hou2019interaction} and a dimensionality-reduction layer is used after average pooling layer, leading to a $512$-D feature vector, analogous to \cite{sun2018beyond,li2018harmonious}.
The model weights are initialized by the parameters of models trained on the ImageNet dataset. StRAttention variant is constructed by replacing the three residual blocks at the last stage by ours where the weights are initialized randomly. Only classification loss (\ie, cross-entropy loss based on identities) is used when training. The normalized (to unit $\ell_2$ norm) feature vectors of query images and gallery images are compared by using the Euclidean distance metric for testing.

When training on Market1501, the parameters of the models from stage 1 to 4 are frozen at the first $8$ epochs that could facilitate convergence. Images are resized to $256\times 128$ and simply augmented by random flipping, cropping, and erasing. We use Adam \cite{kingma2014adam} as the optimizer, where the initial learning rate is set to $3\mathrm{e}{-4}$, and decayed (multiplied) by $0.2$ every $20$ epochs. Batch size is set to $32$ and weight decay is $5\mathrm{e}{-4}$. We train models for 100 epochs with two NVIDIA Tesla P40 GPUs, based on the Pytorch framework. The weight of the divergence loss $\mathcal{L}_d$ added to the objective is set to $1.0$.

\subsection{Face Recognition}
\textbf{Database:} We use a collection of multiple public training datasets \cite{yi2014learning,liu2015deep,chen2014cross,parkhi2015deep} as the medium-size training set, and use VGGFace2 \cite{cao2018vggface2} to show the effectiveness of the proposed method on a larger scale dataset.

For evaluation, we apply the following verification datasets which are typically used for evaluating face models. LFW \cite{article} contains $13,233$ face images from $5,749$ subjects collected from the website. We report the network performance following the standard \textit{unrestricted with labeled outside data} protocol as in \cite{article}. CFP-FP \cite{sengupta2016frontal} dataset aims to evaluate the models when pose variation is high and extreme pose exists. AgeDB-30 \cite{moschoglou2017agedb} contains face images with high age variance. CPLFW \cite{zheng2018cross} and CALFW \cite{zheng2017cross} contain the same identities as LFW while focusing on the evaluation with large pose and age variation, respectively, requiring good generalization of the features extracted from networks.

\textbf{Configuration:} ResNet-50 \cite{he_cvpr2016resnet} is adopted as the backbone network, where conventional global average pooling is replaced by an BN \cite{ioffe_icml2015bn}-Dropout \cite{srivastava2014dropout}-FC-BN module following \cite{deng2019arcface}, and finally produces a $512$-D feature vector. The attention modules are used at the last stage of the
architecture. All the models are trained from scratch. Classification loss (\ie, cross-entropy loss) is used as the objective for training. When evaluating on the test set, the feature vectors extracted from the original images and flipped ones are concatenated and then normalized for comparison. The verification accuracy is conducted with the best threshold on the Euclidean distance metric (in the range of [0,4]) following \cite{wang2018cosface, deng2019arcface}. 

All the models (including baselines and ours) are trained from scratch. Standard SGD with momentum is used for optimization.  Batch size is set to $512$ (on $8$ GPUs, \ie, $64$ per GPU) and weight decay is $5\mathrm{e}{-4}$. For training on the medium-scale dataset, models are trained for $20$ epochs, and the learning rate is initially set to $0.1$ and multiplied by $0.1$ at the $8$-th and $12$-th epochs. Models are trained for $50$ epochs on VGGFace2 dataset \cite{cao2018vggface2}, and the learning rate is initially set to $0.1$ and multiplied by $0.1$ at the $20$-th,$30$-th,$38$-th,$44$-th and $48$-th epochs.

For both training and evaluation sets, images are pre-processed by following standard strategies \cite{wang2018cosface}, \ie, detecting face area and then aligning it to canonical views by performing similarity transformation based on five detected landmarks. Models are trained with center crops (the size is $112\times 112$) of images whose shorter edges are resized to $112$ on the medium-scale dataset. Models trained on VGGFace2 are based on inputs first resized to $224\times 192$. Each pixel is subtracted $127.5$ and divided by $128$ for normalization. Only random horizontal flipping is used as data augmentation during training. The weight of the divergence loss $\mathcal{L}_d$ added to the objective is set to $0.1$.

\subsection{Facial Expression Recognition}
\textbf{Database:} The Facial Expression Recognition 2013 (FER2013) database contains $35,887$ images. The dataset contains $28,709$ training images, $3,589$ validation (public test) images, and another $3,589$ (private) test images. Faces are labeled as any of the seven expressions: “Anger”, “Disgust”,  “Fear”, “Happiness”, “Sadness”,  “Surprise” and  “Neutral”.

\textbf{Configuration:} Images are resized to $44\times 44$ and random horizontal flip is adopted for training. All the models (including baselines and ours) are trained from scratch by using the classification loss. The baseline architecture is a variant of bottleneck residual network, where the kernel size and the stride of the first convolutional layer is set to $3$ and $1$ and followed by BN and ReLU units, max-pooling layer is omitted, and each of the following four stages is comprised of $3$ blocks. We implement the StRA variant by applying the modules at the last stage of the baseline network. The last $1 \times 1$ convolution, which typically fuses feature maps across channels, is omitted in order to facilitate the understanding of behavior among modes.  The weight of the divergence loss is set to $1$. Batch size is set to $128$ (on single GPU). We use SGD with momentum $0.9$ for optimization. Weight decay is set to $5\mathrm{e}{-4}$. The learning rate is initially set to $0.01$ and decayed by $0.9$ every $5$ epochs after $80$ epochs, following \cite{qin2018visual}. Models are trained for $190$ epochs in total.

\begin{table}[b]
\centering
\caption{Ablation studies on module components.}
\label{tab:ablation:module}
{
\renewcommand\arraystretch{1.0}
\begin{tabular}{lp{1.2cm}<{\centering}p{1.8cm}<{\centering}p{1.2cm}<{\centering}p{0.8cm}<{\centering}p{0.8cm}<{\centering}}
\toprule
Model & \multicolumn{3}{c}{Component} & mAP  & Rank1 \\
\cmidrule{2-4}
& Local Attn. & Mode Attn. w/o Interact. & Mode Attn. & & \\
\midrule
ResNet50 &  &  & & 77.1 & 90.6 \\
\midrule
\multirow{3}*{StRAttention} & \checkmark & & & 79.9 & 92.3  \\
& \checkmark & \checkmark & & 83.3 & 93.4 \\
&\checkmark &\checkmark &\checkmark &\bd{84.1} & \bd{93.8} \\
\bottomrule
\end{tabular}
}
\end{table}

\section{Ablation Study on Module Configuration}
We conduct ablation studies on Market1501. The StRAttention block consists of two components, i.e., Local Attention which integrates information over spatially-adjacent regions, and Mode Attention which models the long-range contextual relationships in a node-to-mode manner where mode interaction is exploited to allow mode interactions.

We first assess each component and the results are shown in Table~\ref{tab:ablation:module}. Using our Local Attention can simply yield obvious performance improvement over the baseline ResNet-50, by 2.8\% on mAP and 1.7\% on Rank-1 score. Incorporating Mode Attention without mode interaction can further boost performance. Using the default configuration is able to push the performance further, demonstrating the necessity of all the components in the module. We can conclude that each component plays an important role, and accumulated benefits can be achieved by combining them. 

To validate the effectiveness of the proposed mode attention, we conduct ablation studies by replacing Local Attention with convolutions, group convolutions and SASA \cite{ramachandran2019stand}. Results in Table~\ref{tab:ablation:replace_local} show that replacing the local attention module by classical convolutions or self-attention layers can still obtain superior performance over the counterparts without mode attention, demonstrating the generalizability of the proposed structure regularized paradigm on representation learning. 

\begin{table}[t]
\centering
\caption{Ablation studies on Mode Attention.}
\label{tab:ablation:replace_local}
{
\begin{tabular}{lp{1.8cm}<{\centering}p{1.cm}<{\centering}p{1.cm}<{\centering}p{1.cm}<{\centering}}
\toprule
Method & mAP & Rank1 & Params & FLOPs \\
\midrule
Conv & 77.1 & 90.6 & 24.6M   & 4.05G \\
SASA \cite{ramachandran2019stand} & 79.5 & 92.3 & 17.8M   &3.19G \\
\midrule
{Conv + Mode} & 82.1 & 93.2 & 24.6M & 4.06G \\
{Group Conv + Mode} & 82.8 & 93.3 & 18.4M & 3.26G \\
{SASA + Mode} & 83.0 & 93.6 & 17.8M & 3.19G \\
{Local + Mode} (ours) & 84.1 & 93.8 & 17.6M & 3.17G \\
\bottomrule
\end{tabular}
}
\end{table}

\bibliography{strattention_2020}
\bibliographystyle{plain}
\end{document}